\pdfoutput=1

\documentclass[11pt]{article}

\usepackage{ACL2023}
\RequirePackage[inline]{enumitem} \setlist{nosep}

\newcommand{\modelname}{\texttt{X-InSTA}}
\newcommand{\fullname}{Cross-lingual {\bf In}-context {\bf S}ource-{\bf T}arget {\bf A}lignment}

\newcommand\blfootnote[1]{%
  \begingroup
  \renewcommand\thefootnote{}\footnote{#1}%
  \addtocounter{footnote}{-1}%
  \endgroup
}

\usepackage{ragged2e}
\usepackage{amsmath}
\usepackage{diagbox}
\DeclareMathOperator*{\argmax}{argmax}

\usepackage{times}
\usepackage{latexsym}

\usepackage[T1]{fontenc}

\usepackage[utf8]{inputenc}

\usepackage{microtype}

\usepackage{inconsolata}

\usepackage{color, colortbl}
\definecolor{Gray}{gray}{0.9}
\usepackage{multirow,adjustbox}
\newcolumntype{M}[1]{>{\centering\arraybackslash}m{#1}}

\usepackage[ruled,vlined]{algorithm2e}

\usepackage{bm,amsmath,amssymb,fancyhdr,graphicx}
\usepackage[disable=True]{todonotes}
\usepackage{hyperref}

\title{Multilingual LLMs are Better Cross-lingual \\In-context Learners with Alignment}

\author{Eshaan Tanwar \\
   DTU, India \\
   \texttt{eshaantanwar2000@gmail.com} \\\And
   Subhabrata Dutta \\
   IIT Delhi, India \\
   \texttt{subha0009@gmail.com} \\ \AND
   Manish Borthakur \\
   IIT Delhi, India \\
 \texttt{mt6190493@iitd.ac.in} \\ \And
   Tanmoy Chakraborty \\
   IIT Delhi, India \\
   \texttt{tanchak@iitd.ac.in} \\}

\begin{document}
\maketitle
\begin{abstract}
In-context learning (ICL) unfolds as large language models become capable of inferring test labels conditioned on a few labeled samples without any gradient update. ICL-enabled large language models provide a promising step forward toward bypassing recurrent annotation costs in a low-resource setting. Yet, only a handful of past studies have explored ICL in a {\em cross-lingual setting}, in which the need for transferring label-knowledge from a high-resource language to a low-resource one is immensely crucial. To bridge the gap, we provide the first in-depth analysis of ICL for cross-lingual text classification. We find that the prevalent mode of selecting random input-label pairs to construct the prompt-context is severely limited in the case of cross-lingual ICL, primarily due to the lack of alignment in the input as well as the output spaces. To mitigate this, we propose a novel prompt construction strategy -- \fullname\ (\modelname). With an injected coherence in the semantics of the input examples and a task-based alignment across the source and target languages, \modelname\ is able to outperform random prompt selection by a large margin across three different tasks using 44 different cross-lingual pairs. 
 \blfootnote{ET and SD contributed equally. ET and SD designed the experiments. ET and MB ran the experiments. SD and TC wrote the paper. TC mentored the project.} 
\end{abstract}

\section{Introduction}
\label{sec:intro}

The emergence of large-scale, pretrained, Transformer-based language models (LLMs) has marked the commencement of an avant-garde era in NLP. Departing from the traditional methods of neural language learning with temporally separated training-testing phases for downstream tasks, 
\begin{figure*}[h]
\centering
\includegraphics[width=16cm]{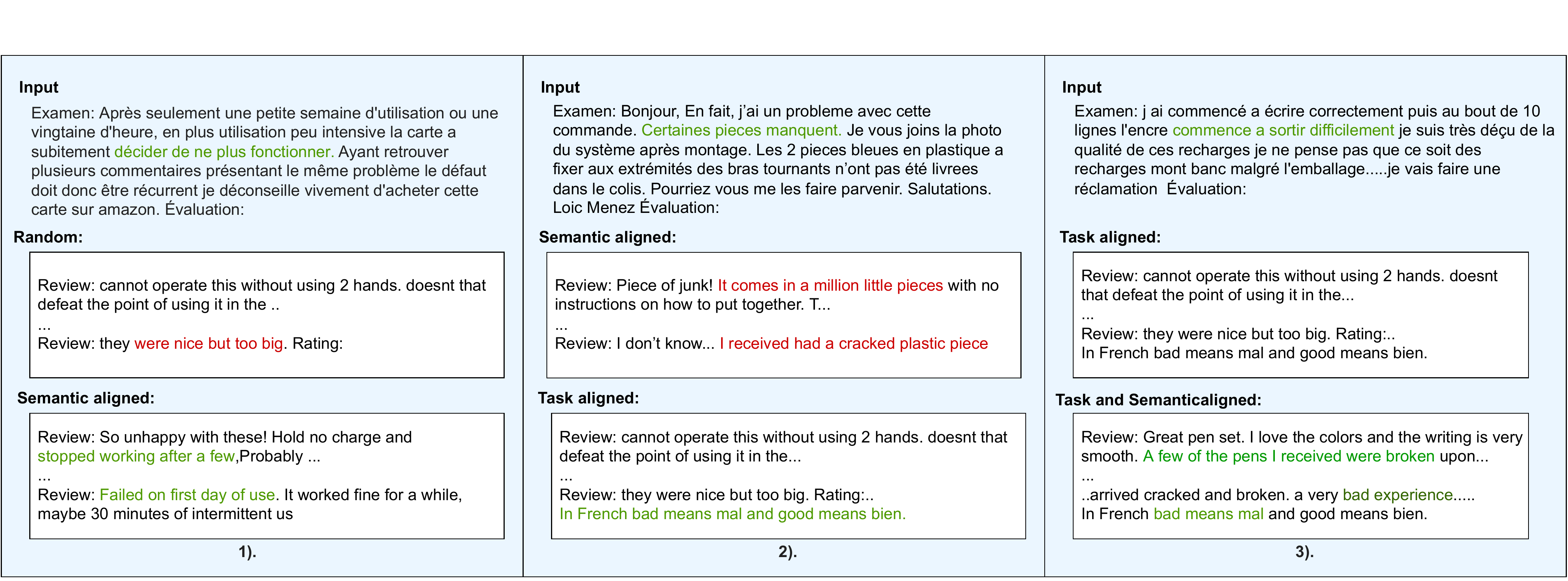}
\caption{Working example of different ICL prompts explored in this work. In example \#1,  randomly selecting the prompt examples fails as it prompts irrelevant contradictions, whereas semantic alignment succeeds as it makes the context with similar reviews. In example \#2, semantic alignment fails; it extracts demonstrations about `multiple pieces', but these are not helpful for the LLM, whereas a simple task aligner works. In the last example, it is a combination of semantic and task alignments that works.}
\label{fig:x-insta-example}
\end{figure*}
pretrained LLMs have shown the ability to infer labels from test inputs conditioned on the training data within a single pass. This is known as \textit{In-context learning} -- an LLM is prompted with a few input-output pairs from the training data (commonly referred to as \textit{demonstrations}) followed by the test input; for generative tasks (summarization, text-to-code, chain-of-thought reasoning, etc.) the LLM is then required to produce an output; for classification tasks, the probabilities of the next tokens predicted by the LLM are mapped to the label space. All of this is done without updating the parameters of the LLM. In-context learning is particularly promising for two different aspects. Firstly, it reduces the need for task-specific training data, and thus, the cost of human annotation. Secondly, while the LLM was trained in a compute-intensive environment, the removal of the need for task-specific gradient-based weight updates can significantly reduce the carbon footprint of automated NLP/NLU since the inference-time compute-necessity is orders of magnitude smaller than that of the training/finetuning phases. Multiple recent advancements have been proposed to optimize the ICL ability of the LLMs~\cite{lin2021few, chowdhery2022palm, liu-etal-2022-makes, zhang2021differentiable}.

{\bf Challenges in cross-lingual ICL:} Given that there is an order-of-magnitude discrepancy in the availability of annotated data in a high-resource language vs. a low-resource one, the ability to learn from the high-resource source context to solve tasks in low-resource targets sounds enticing. Yet, the application of ICL in a cross-lingual setting remains largely unexplored. Previous attempts at multilingual ICL~\citep{zhang2021differentiable, winata-etal-2021-language} use randomly selected input-label pairs to construct the prompt-context. This limits the ability of an LLM to infer from the context. As \citet{icl-bayesian} suggested, ICL emerges as the ability to infer target labels from the pretraining distribution conditioned upon the context; each input-label pair in the prompt-context are, in turn, sampled from the prompt token distribution. Theoretically, the expected prediction error decreases as the number of examples in the prompt increases. However, such {\em infinitely long} prompts are practically infeasible to attain. \citet{icl-bayesian} imposed that a distinguishability of the prompt-concept, shared across the prompt-examples, from all other possible concepts is essential for an optimal predictor. A random sampling of prompt examples is unlikely to construct a prompt with distinguishable concepts. Furthermore, given $(x_i, y_i)$ and $(x_{i+1}, y_{i+1})$ as two consecutive input-label pairs in the prompt-context, the transition probability from $y_i$ to $x_{i+1}$ is a low-probability one under the pretraining distribution~\cite{icl-bayesian}. The transition becomes even more improbable if we are to simply append a test example to the prompt-context of a different language. Consider the following example of ICL prompting for cross-lingual sentiment classification:
\begin{center}
\small
\begin{tabular}{|ll|l|}
\hline
    1. & That movie was good. & Positive\\\hline
    2. & Depression is the new pandemic. & Negative\\\hline
    3. & Ella lo está haciendo bien & ?\\
\hline
\end{tabular}
\end{center}
The text segments are concatenated from left-to-right and top-to-bottom; therefore, two English input-label pairs are followed by a Spanish test input. There are irremovable, token-level low-probability transitions from the labels to the next input sentences. On top of this, we have three completely unrelated sentences juxtaposed together with an abrupt change in language. Intuitively, it is less likely for an LLM to be able to map the third input to its correct label, \textit{positiva} (positive in Spanish) following the very much convoluted patterns presented in English. 

{\bf 
Proposed approach:} We seek to develop prompt-design strategies for ICL in a cross-lingual setting that can overcome the foregoing challenges. A two-way alignment of the source and target examples is proposed. We start with injecting semantic coherence into the prompt-context by selecting similar examples; this aligns the labeled demonstrations as well as the test inputs to share a set of common concepts. Next, we seek to enforce an alignment of task-level signals across languages. We introduce manually-designed task-specific mappings from the source language to the target language, thereby providing the LLM with a `natural' transition from the former to the latter. Together, these two approaches constitute our proposed prompts-selection strategy,  \modelname\ (\fullname, see Figure~\ref{fig:x-insta-example} for working examples). \modelname\ shows a staggering $\bf 18\%$ relative improvement over random prompt selection averaged across three different text classification tasks in multiple different languages with English being the source language. Careful perturbations to these alignment methods disclose the importance of label space structure induced by LLMs for cross-lingual ICL.

Our contributions are summarized below\footnote{Code available at \url{https://github.com/EshaanT/X-InSTA}}:
\begin{description}[leftmargin=0cm]
\item[1.] We propose \modelname, a novel method of aligning prompt examples in a cross-lingual scenario. To the best of our knowledge, {\em this is the first attempt to push prompt design techniques for ICL in cross-lingual settings beyond the trivial strategy of random example selection}.
\item[2.] We present the first, in-depth analysis of the role of semantic similarity between prompt examples for cross-lingual ICL.
\item[3.] A novel concept of task-based prompt alignment is presented. We show its efficacy with 44 different source-target language pairs and empirically relate this to the underlying structures of multilingual representations of the LLM.

\end{description}










\section{Prompting Techniques}

In this section, we lay out a step-by-step approach to aligning semantic coherence and task-based signals across source-target examples for ICL prompts.

\subsection{Prelimineries}
Let $D_s = \{(x^i_s,y_s^i)\}_{i}$ be a monolingual labeled dataset in language $s$, realized as a collection of input examples and their labels, $x^i_s\in X_s$ and $y_s^i\in Y_s$, respectively. Here $Y_s$ is the natural language label space in language $s$. We have another collection of input examples, $D_t=\{x^i_t\}_i$, with examples in language $t$. One can define a cross-lingual text classification task with source and target languages being $s$ and $t$ in the following manner. First, we select $k$ input-label pairs from $D_s$ to construct the prompt-context, $C$:
\begin{equation}
    C= x_{s}^1\oplus y_{s}^1\oplus[sep]\oplus \cdots x_{s}^k\oplus y_{s}^k
\end{equation}
where $[sep]$ denotes a separator token (e.g., newlines), and $\oplus$ denotes the concatenation operator. The problem of in-context prediction then translates to inferring the label $y_t\in Y_t$, where $Y_t$ is the natural language label space in language $t$ corresponding to the test input $x_t\in D_t$ conditioned on the prompt-context $C$, as follows:
$$y_t=\argmax_{y\in Y_t}p(y|C\oplus x_t)$$
i.e., we select the maximum probability label in the target label space generated by the model as the token next to the test input $x_t$ appended to the context $C$. The source and target label spaces, $Y_s$ and $Y_t$, share a one-to-one mapping among each other in terms of translation from $s$ to $t$.



One of the most widely-used methods of constructing the context $C$, which we will henceforth call {\bf random prompting}, is to randomly select $(x^i_s, y^i_s)$ from $D_s$ and concatenate together. We explore this method in our analysis, and it serves as a baseline for our experiments.

\subsection{Semantic Alignment}\label{sec:sim-promp}

\citet{chang2022geometry} showed that  multilingual models encode these languages in a shared embedding space, while still preserving several language-sensitive semantic information. Despite the language difference between source and target inputs, $x_s$ and $x_t$, it is then likely that their semantic similarities will be reflected in their hidden representations constructed by  LLM. Therefore, we hypothesize that choosing semantically similar examples to construct the prompt-context would help the model do in-context inference. That is, if ${\bf e}_t$ is the embedding of the target and ${\bf e}_s$ that of the source, the higher the similarity score between them, the better sentence $x_s$ will serve as a demonstration for the target sentence $x_t$.

Inspired by \citet{liu-etal-2022-makes}, we extract prompt examples directly dependent on the test input distribution. Here we utilize multilingual sentence-transformers \cite{reimers-2020-multilingual-sentence-bert} to extract the sentence embedding of the test input $x_{t} \in D_t$ and the source inputs $X_s$.  Based on the cosine similarity between the target input $x_{t}^{j}$ and source inputs $x^j_{s}\in X_s$, we then extract the top $k$ demonstrations (see Algorithm~\ref{alg:sim-prompt}). While the target input and the demonstration differ in language, we hypothesize that by pairing semantically similar context demonstration and input sentence, the LLM would be able to improve its reasoning ability and subsequently, the final task performance (see Table~\ref{tab:prompt-demo-MARC} in Appendix \ref{sec:appendix} for examples of such aligned demonstrations).

\begin{algorithm}\small
\SetAlgoLined
\caption{Semantic Alignment}\label{alg:sim-prompt}
\textbf{Input: }An unlabeled target sentence $x_t$, source data $D_s$, multilingual sentence encoder, $\theta$, and number of samples to extract $k$.

\textbf{Procedure: }
$\mathbf{e_{t}}\gets\theta(x_t)$\\
\For{$x^s\in D_s$}{
    $\mathbf{e_{s}^i}\gets\theta(x_{s}^i)$
    
    $s_i\gets\frac{\mathbf{e_{t}.e_{s}^i}}{||e^{t}||_2||e_s^{i}||_2}$}
    
Select top $k$ sentences based on $s_i$ 

$C\gets x_{s}^1\oplus y_{s}^1\oplus[sep]\oplus \cdots x_{s}^k\oplus y_{s}^k$

$y_t=\argmax_{y\in Y_t}p(y|C\oplus x_t)$
\end{algorithm}


\subsection{Task-based Alignment}\label{sec:log-promp}

Despite the semantic coherence enforced within the prompt-context via the previously mentioned method, the source and target label spaces, $Y_s$ and $Y_t$, remain superficially disconnected. 
For fine-tuning, techniques like meta-learning \cite{nooralahzadeh2020zero}, and adapters \cite{parovic-etal-2022-bad} have been used to bridge this gap. For in-context prompting in which context matters the most, we propose to do so by adding a manually designed statement that gives the LLM task-specific information like target language and target label space. 

{Task-based alignment} is done by appending a manually-designed statement, called {\em task aligner} to context. This aligner is supposed to inform the LLM about the mapping from the source label space $Y_s$ to the target label space $Y_t$. We do task alignment by first manually creating $D_l=\{L_{s,t}\}$ for a given task and source-target language pairs $s$ and $t$ as a collection of statements in the source language that emphasizes what the target label and language are. For example, when the source is English and the target is Spanish, ``In Española bad means malo and good means bueno'' will be the said task aligner that gives the information that the target language is Española (Spanish) and the target labels are {\em malo} and {\em bueno} ({\em bad} and {\em good}, respectively). Next, we construct the prompt-context by randomly selecting $k$ source language examples, followed by the task aligner from this source-target pair from $D_l$ (see Algorithm~\ref{alg:log-prompt}). For more examples of task-aligned prompt design, please refer to Tables~\ref{tab:prompt-demo-MARC} and  \ref{tab:task-aligner-demo} in Appendix \ref{sec:appendix}.

  \begin{table}[ht]
\small
\setlength{\tabcolsep}{3.5pt}
\centering
\begin{tabular}{c|cccccc}
 \hline
 \backslashbox{ SRC}{ TAR}&de&en&es&fr&ja&zh\\
 \hline
 \rowcolor{Gray}\multicolumn{7}{c}{Random Prompting}\\
 \hline
 de&$-$&$0.446$&$0.517$&$0.547$&$0.454$&$0.413$\\
 en &$\textbf{0.380}$&$-$&$0.761$&$\textbf{0.663}$&$\textbf{0.526}$&$0.362$\\
 es&$0.339$&$0.696$&$-$&$0.563$&$0.519$&$0.445$\\
 fr&$0.340$&$0.692$&$\textbf{0.864}$&$-$&$0.479$&$0.410$\\
 ja&$0.333$&$\textbf{0.701}$&$0.678$&$0.612$&$-$&$\textbf{0.678}$\\
 zh&$0.333$&$0.632$&$0.836$&$0.402$&$0.521$&$-$\\
 \hline
 AVG&$0.345$&$0.633$&$0.731$&$0.557$&$0.499$&$0.462$\\
 \hline
 \rowcolor{Gray}\multicolumn{7}{c}{Semantic Alignment}\\
  \hline
 de&$-$&$0.6$&$0.552$&$0.679$&$0.559$&$0.483$\\
 en &$\textbf{0.458}$&$-$&$0.783$&$\textbf{0.762}$&$0.608$&$0.450$\\
 es&$0.377$&$\textbf{0.771}$&$-$&$0.740$&$0.643$&$0.568$\\
 fr&$0.376$&$0.752$&$\textbf{0.879}$&$-$&$0.565$&$0.589$\\
 ja&$0.333$&$0.754$&$0.733$&$0.690$&$-$&$\textbf{0.697}$\\
 zh&$0.333$&$0.682$&$0.839$&$0.536$&$\textbf{0.675}$&$-$\\
 \hline
 AVG&$0.375$&$0.713$&$0.757$&$0.681$&$0.610$&$0.557$\\
 \hline
 \rowcolor{Gray}\multicolumn{7}{c}{Task-based Alignment}\\
  \hline
 de&$-$&$0.567$&$0.701$&$0.768$&$0.645$&$0.333$\\
 en &$\textbf{0.355}$&$-$&$\textbf{0.888}$&$0.826$&$0.727$&$0.333$\\
 es&$0.334$&$0.784$&$-$&$0.806$&$\textbf{0.779}$&$0.333$\\
 fr&$0.336$&$0.783$&$0.827$&$-$&$0.766$&$0.333$\\
 ja&$0.333$&$\textbf{0.796}$&$0.864$&$\textbf{0.847}$&$-$&$\textbf{0.345}$\\
 zh&$0.333$&$0.682$&$0.872$&$0.543$&$0.734$&$-$\\
 \hline
 AVG&$0.338$&$0.722$&$0.830$&$0.758$&$0.730$&$0.335$\\
 \hline
 \rowcolor{Gray}\multicolumn{7}{c}{\modelname}\\
 \hline
 de&$-$&$0.721$&$0.756$&$0.847$&$0.760$&$0.333$\\
 en &$\textbf{0.382}$&$-$&$0.891$&$0.858$&$0.783$&$0.335$\\
 es&$0.348$&$\textbf{0.857}$&$-$&$\textbf{0.875}$&$\textbf{0.851}$&$0.334$\\
 fr&$0.356$&$0.849$&$\textbf{0.906}$&$-$&$0.825$&$0.336$\\
 ja&$0.333$&$0.832$&$0.890$&$0.845$&$-$&$\textbf{0.348}$\\
 zh&$0.333$&$0.717$&$0.883$&$0.684$&$0.809$&$-$\\
 \hline
 AVG&$0.350$&$0.795$&$0.865$&$0.822$&$0.805$&$0.337$\\\hline

\end{tabular}
\caption{Macro-F1 scores for different prompting techniques on the MARC dataset (source and target languages are abbreviated as { SRC} and { TAR}, respectively). Improvement across all six languages can be observed once we introduce semantic alignment. \modelname\ outperforms rest of the methods on 4 out of 6 languages.}
 \label{tab:main-results-MARC}
\end{table}

\begin{table}[h]
\centering
\small
\begin{tabular}{c|cccc}
 \hline
 \backslashbox{ Source}{\ Target}&de&en&fr&ja
 \\
  \hline
 \rowcolor{Gray}\multicolumn{5}{c}{Random Prompting}\\
 \hline
 de&$-$&$0.517$&$\textbf{0.597}$&$0.618$\\
 en &$\textbf{0.682}$&$-$&$0.412$&$0.609$\\
 fr&$0.545$&$\textbf{0.694}$&$-$&$\textbf{0.666}$\\
 ja&$0.344$&$0.595$&$0.475$&$-$\\
 \hline
 AVG&$0.524$&$0.602$&$0.495$&$0.631$\\
 \hline
 \rowcolor{Gray}\multicolumn{5}{c}{Semantic Alignment}\\
 \hline
  de&$-$&$0.502$&$\textbf{0.643}$&$0.657$\\
 en &$\textbf{0.677}$&$-$&$0.505$&$0.691$\\
 fr&$0.572$&$\textbf{0.746}$&$-$&$\textbf{0.743}$\\
 ja&$0.344$&$0.617$&$0.481$&$-$\\
 \hline
 AVG&$0.531$&$0.621$&$0.543$&$0.697$\\
 \hline
 \rowcolor{Gray}\multicolumn{5}{c}{Task Alignment}\\
  \hline
 de&$-$&$0.618$&$\textbf{0.741}$&$\textbf{0.753}$\\
 en &$\textbf{0.620}$&$-$&$0.696$&$0.752$\\
 fr&$0.511$&$\textbf{0.782}$&$-$&$0.824$\\
 ja&$0.339$&$0.658$&$0.697$&$-$\\
 \hline
 AVG&$0.490$&$0.686$&$0.711$&$0.776$\\
\hline
  \rowcolor{Gray}\multicolumn{5}{c}{\modelname}\\
 \hline
  de&$-$&$0.622$&$\textbf{0.788}$&$0.779$\\
 en &$\textbf{0.588}$&$-$&$0.778$&$0.794$\\
 fr&$0.524$&$\textbf{0.821}$&$-$&$\textbf{0.834}$\\
 ja&$0.339$&$0.701$&$0.705$&$-$\\
 \hline
 AVG&$0.483$&$0.715$&$0.757$&$0.803$\\\hline

\end{tabular}
\caption{Macro F1 scores on the CLS dataset.}
 \label{tab:main-results-cls}
\end{table}

\begin{table}[ht]
\small
\begin{center}
\begin{tabular}{ c|cc}
 \hline
 \backslashbox{ Source}{ Target}&es&en \\
  \hline
 \rowcolor{Gray}\multicolumn{3}{c}{Random Prompting}\\
 \hline
 es&$-$&$0.274$\\
 en&$0.435$&$-$\\
 \hline
 AVG&$0.435$&$0.274$\\
 \hline
\rowcolor{Gray}\multicolumn{3}{c}{Semantic Alignment}\\
\hline
es&$-$&$\textbf{0.284}$\\
en&$0.493$&$-$\\
\hline
AVG&$0.493$&$0.284$\\
 \hline
\rowcolor{Gray}\multicolumn{3}{c}{Task Alignment}\\
\hline
 es&$-$&$0.269$\\
 en&$0.499$&$-$\\
 \hline
AVG&$0.499$&$0.269$\\
 \hline
\rowcolor{Gray}\multicolumn{3}{c}{\modelname}\\
\hline
 es&$-$&$0.269$\\
 en&$\textbf{0.542}$&$-$\\
 \hline
 AVG&$0.542$&$0.269$\\
\hline

\end{tabular}
\end{center}
\caption{Macro F1 scores on the HatEval dataset.}
 \label{tab:main-results-hateval}
\end{table}

\begin{table*}[h]
\small
\centering
\begin{tabular}{l|cccccc}
 \hline
 \backslashbox{ Setup}{ Target language}&de&en&es&fr&ja&zh\\
 \hline
{Random prompt}&0.345&0.633&0.731&0.557&0.499&0.462\\
{Uniform label space}&\textbf{0.441}&0.570&0.493&0.414&0.483&\textbf{0.594}\\
{Task alignment by language information only}&0.346&0.645&0.733&0.575&0.543&0.508\\
{Task alignment via third language}&0.345&0.687&0.755&0.673&0.601&0.423\\
{Incorrect task alignment}&0.338&0.665&0.787&0.647&0.544&0.339\\
{Task Alignment}&0.338&\textbf{0.722}&\textbf{0.830}&\textbf{0.758}&\textbf{0.730}&0.335\\
\hline
\end{tabular}
\caption{Understanding how task alignment works. Average F1-Macro across all source-target pairs on MARC.}
 \label{tab:main-why-logic}
\end{table*}


\begin{algorithm}\small
\SetAlgoLined
\caption{Task Alignment}\label{alg:log-prompt}
\textbf{Input: }An unlabeled target sentence $x_t$, source dataset $D_s$, aligner $L_{s,t}$ and number of samples to extract $k$.

\textbf{Procedure: }
Randomly select $k$ sentences from $D_s$

$C\gets x_{s}^1\oplus y_{s}^1\oplus[sep]\oplus \cdots x_{s}^k\oplus y_{s}^k$

$C\gets C\oplus L_{s,t}$

$y_t=\argmax_{y\in Y_t}p(y|C\oplus x_t)$
\end{algorithm}

\subsection{\modelname} 

We finally move on to our proposed method \modelname\ that combines  semantic alignment with the task-based one. It first selects source examples from $D_s$ with top-$k$ similarity scores as mentioned in Section~\ref{sec:sim-promp}. Additionally, we select task-aligners from $D_l$ depending on the source and target languages and the task. Finally, we construct the prompt context by concatenating the selected examples followed by the task-aligner. The final label inference can be described as
$$y_t = \argmax_{y\in Y_t} p(y|x^1_s\oplus y^1_s\cdots x^k_s\oplus y^k_s \oplus L_{s,t}\oplus x_t)$$
where $\operatorname{sim}(x^i_s, x_t)\geq \operatorname{sim}(x^{i+1}_s, x_t)$, and $L_{s,t}\in D_l$ is the task aligner for source and target languages $s$ and $t$, respectively for the given task.

\todo{have u even mentioned the abbrev of languages -- de, en etc.}

\begin{table}[h]
\small
\centering
\begin{tabular}{M{2.5cm}||M{0.55cm}M{0.55cm}M{0.55cm}M{0.55cm}}
 \hline
 \backslashbox{  Setup}{  Target}&de&en&fr&ja
 \\
  \hline
  \multicolumn{1}{l||}{Random }& 0.524&0.602&0.495&0.631\\
  \multicolumn{1}{l||}{Non-Semantic}&0.531&0.561&0.453&0.515\\
  \multicolumn{1}{l||}{Semantic}&0.531&0.621&0.543&0.697\\
 \hline

\end{tabular}
\caption{Dissecting the role of semantic alignment; we present macro-F1 scores corresponding to different prompting techniques on the CLS dataset for each source language averaged over all target languages.}
 \label{tab:sim-dis-sim}
\end{table}

\section{Results and Analysis}
We experiment on three datasets -- Multilingual Amazon Reviews Corpus (MARC) \cite{keung-etal-2020-multilingual}, Cross-language sentiment classification (CLS) \cite{prettenhofer-stein-2010-cross}, and HatEval \cite{basile-etal-2019-semeval}, spanning over twelve language-task pairs and totalling $44$ cross-lingual setups (refer to Appendix~\ref{app:data-detail} for further description of the datasets). The results on MARC, CLS and HatEval are shown in Tables \ref{tab:main-results-MARC},  \ref{tab:main-results-cls}, and \ref{tab:main-results-hateval}, respectively. For our main experiments, we make use of XGLM~\cite{lin2021few} 7.5 billion variant. \textcolor{black}{We experiment with various models with random prompting and select XGLM 7.5B for its performance superiority on various tasks (refer to Table~\ref{tab:model-varientd} in Appendix~\ref{app:model-variant})}. 
For further details on the experimental setup, please refer to Appendix~\ref{app:hyperparam} and Table~\ref{tab:langs_iso} for the language abbreviations used.

\subsection{Comparing Alignment Techniques}

\textbf{Semantic  Alignment: } The improvement introduced by semantic alignment of the prompt-context over randomly-selected source examples is eminent in Tables \ref{tab:main-results-MARC}, \ref{tab:main-results-cls}, and \ref{tab:main-results-hateval}. On the MARC dataset, we  observe a 14\% improvement in macro F1 scores averaged across different languages. This observation is consistent across all target-source pairs on other datasets as well --- a gain of 10\% on Hateval, and 6\% on CLS. 
This improvement over random example selection is consistent across all language pairs (except English-to-German in CLS) considered in this experiment. This is particularly noteworthy and one might lead to the conclusion that dynamically selecting prompt examples based on semantic similarity aligns the LLM to become a better in-context learner irrespective of the task and the languages.

\textbf{Task-based Alignment: } Just by adding a task aligner, we not only outperform random prompts but also bring substantial improvements for similarity prompting, even though it is not dynamically varying with input sentences. The improvement is 18\% in CLS, 8\% in HatEval, and 15\% in MARC, in terms of macro F1 scores averaged over different language pairs.

However, some languages like German in MARC and English in HatEval produce near-random predictions in all the set-ups we experimented with. This might be due to the model's inability to perform ICL on these tasks in a cross-lingual manner for these languages. Previous studies observed such phenomena in monolingual ICL \cite{webson-pavlick-2022-prompt,lin2021few}; cross-lingual ICL has its added nuances that make it even more difficult. 

We also see a performance drop in the case of Mandarin in MARC (Table~\ref{tab:main-results-MARC}) while adding a task aligner. We investigate the performance drop and near-random results of German further. 

\textbf{\modelname:} This prompting mechanism inherits both the benefits of semantic and task-based prompting, hence giving the best results in most language pairs. But similar to task-based alignment, \modelname\ also performs badly on some target languages. The improvement is 23\% on MARC, 22\% on CLS, and 14\% on HatEval.
We also note that no specific language can be used as the best source language.

\subsection{Why does Task Alignment Work?}\label{sub-sec:why-logic-works} Next, we seek to validate the performance boost achieved via task-based aligners along with an attempt to explain the drop in performance with Mandarin and German. We vary the task aligner and note its effect on the output. We do so in five different variations along with the original method (see Table~\ref{tab:task-aligner-demo} in Appendix~\ref{sec:appendix} for detailed examples of each scenario):
\begin{enumerate}
    \item \textbf{No aligner prompt added: } Same as random prompting.
    \item \textbf{Making the label space uniform: } Across all source-target setups, we set the source-label distribution as output for the target too, reducing the need for task alignment.
    \item \textbf{Only language information: } Only giving the language information to  LLM, without providing any further label information. An example of such an aligner would be `The following post is in \textit{French} language', in a case when the source is English, and the target is French.
    \item \textbf{Providing aligner but of a third unrelated language: } We set the aligner of a third language. For example `In Spanish bad means \textit{malo} and good means \textit{bueno}.', in a case when the source is English and the target is French.
    \item \textbf{Incorrect aligner: } Making the aligner incorrect corresponding to the label space. For example `In French bad means \textit{bien} and good means \textit{mal}.', in a case when the source is English and the target is French.
\end{enumerate}

\textbf{It's all about the label information: } In Table \ref{tab:main-why-logic}, we note the importance of label space information. Providing the model with language information does improve the performance; however,  the improvement is minuscule compared to the improvement achieved via task aligners. This label information, even when of an unrelated third language, still helps the model predict better. This might be due to the fact that the model looks more rigorously at label space for inference. 
Therefore, this showcases the importance of labelling information while going cross-lingual.

\textbf{Why drop in some languages? } It is noteworthy that in Table \ref{tab:main-why-logic}, the task aligner works best for all target languages except for German and Mandarin. Both of these languages give the best results in uniform label space, i.e., when $y_t$ is made the same as $y_s$. This points to the inability of the LLM to align the label space of different source languages to these target languages. In making the label space uniform, we lose certain language-specific signals, but this may also be seen as a way of reducing task alignment. Only for German and Mandarin do we see this trade-off as beneficial; in all other cases, the loss of language-specific features of $y_t$ leads to a drop in performance.

\subsection{Role of semantic alignment}
To understand the role of semantic alignment, we ran an experiment in which instead of choosing $k$ nearest neighbor of $x_t$, we chose the most dissimilar sentences.  Table \ref{tab:sim-dis-sim} shows that there is a sharp decrease in performance as compared to random prompting for all languages, with  German as an exception. The average fall is 8\% whereas using semantic alignment gives a gain of 10\% w.r.t. random prompting.



\begin{table*}[h]
\small
\centering
\adjustbox{max width=1.2\linewidth}{
\begin{tabular}{l|ccccc|ccc|c}
 \hline
 \multirow{2}{*}{\backslashbox{ Setup}{ Target}} &\multicolumn{5}{c|}{MARC}&\multicolumn{3}{c|}{CLS}&\multicolumn{1}{c}{HatEval}\\ 
 &de&es&fr&ja&zh&de&fr&ja&es
 \\
 \hline
{Random prompting}&0.380&0.761&0.663&0.526&0.362&0.682&0.412&0.609&0.435\\
{ Semantic alignment}&0.458&0.783&0.762&0.608&\textbf{0.450}&0.677&0.505&0.691&0.493\\
{Task-based alignment}&0.355&\textbf{0.888}&\textbf{0.826}&\textbf{0.727}&0.333&0.620&\textbf{0.696}&\textbf{0.752}&\textbf{0.499}\\
 {Automated aligner}&\textbf{0.531}&0.792&0.699&0.599&0.350&{0.721}&0.430&0.610&0.438\\\hline
\end{tabular}
}
\caption{Comparing the performance of automated aligners generated by mT5 with the rest of the methods in terms of macro-F1. We use English as the source language for all three tasks in this experiment.}

\label{tab:auto-logic}
\end{table*}
\subsection{Automated aligner generation}
We also expand our analyses to automatically generate the aligner using mT5 \cite{xue-etal-2021-mt5}. It is trained using a span generation task using sentences like `Paris <MASK> France'. The mT5 model is trained to fill the mask token by generating spans like `is capital of'. In our usage, mT5 will fill the <MASK> between the input target test $x_t$, and prompt context $C$ in the source language  to align the semantics of both. We summarize our procedure for automatic alignment generation in Algorithm~\ref{alg:auto-log-prompt}.
\begin{algorithm}\small
\SetAlgoLined
\caption{Task Alignment}\label{alg:auto-log-prompt}
\textbf{Input: }An unlabeled target sentence $x_t$, source data set $D_s$, multilingual-T5, $mT5$, multilingual LLM, $M$ and number of samples to extract $k$.

\textbf{Procedure: }
Randomly select $k$ sentences from $D_s$

$C\gets x_{s}^1\oplus y_{s}^1\oplus[sep]\oplus \cdots x_{s}^k\oplus y_{s}^k$

$L\gets mT5(C\oplus [MASK] \oplus x_t)$, where L is the generated span

$C\gets C\oplus L$

$y_t=\argmax_{y\in Y_t}p(y|C\oplus x_t)$
\end{algorithm}

\begin{table*}[ht]
\small
\begin{center}
\scalebox{0.9}{
\begin{tabular}{ |m{0.1cm}|m{10cm}| m{5cm}| m{0.85cm}|}
 \hline
 \#&{\bf Prompt}&{\bf Input}&{\bf Output} \\
  \hline
1.&Post: @anne\_theriault @hels Shut up cunt bitch whore Is hatefull? yes

Post: @okaayythen bitch ass whore ass motherfucker Is hatefull? yes

Post: @KoiyaThePerv bitch ass hoe Is hatefull? yes

Post: @nonracialnigger Wow shut up you whore Is hatefull? yes

In Spanish no hate means no odio and yes means si odio & correo: @Viplajo @blondeputicorn cállate era una historia de amor hijo de puta Es odioso?  & \textcolor{red}{yes} \\
\hline
2.&
correo: @diegovftbau REENCUENTRO LA PUTA MADRE, DISFRUTALO MUCHO, LO RE MERECES AMIGA Es odioso? no 

correo: @AlejoIgoa @igoaftbau BIAN LA PUTA MADRE TE LO MERECES MAS QUE A NADIE Es odioso? no

correo: @ansecas CÁLLATE LESBIANA HIJA DE PUTA QUE ESTÁ AHÍ TÚ HIJO BORT Es odioso? si

correo: @Isranutrof CALLATE PUTA! MIS JOJOS VUELVEEEEN Es odioso? si 

En English no odio significa no hate y si odio significa yes hate

& Post: Arrest them and DEPORT THEM ALL!!!!\#BuildThatWall Is hateful? & \textcolor{red}{no} \\
\hline
3.&</s> Examen: Bravo à Nicolas Jacquette pour avoir réussi à quitter l'enfer d'une secte qui met au pinacle le sacrifice de la vie des siens...TO LONG CONTEXT....bon signifie good & Review: ﻿In the end, it appears ......\textbf{THE POST IS TOO LONG}..... the problem than for the individual transgressions of certain priests Rating: & \textcolor{red}{good} \\
\hline
4.&</s> 
Review: The mice loved them \& are holes in the bags to get to the inside product.  Rating: bad

Review: Product bag was smashed and bag was spilled out in box.  Rating: bad

Review: The product came in and the spoons are already cracked and broken.  Rating: bad

Review: Item received was broken, with product leaking out and all over the jar.  Rating: bad

</s> In French bad means mal and good means bien.</s>& Examen: Produit bien reçu mais pastilles a l'intérieur des sachets en miettes et un sachet craqué.  Évaluation: & \textcolor{red}{bien} \\
\hline
\end{tabular}
}
\caption{Error analysis of \modelname. Four examples represent the major error characteristics (discussed in Section~\ref{subsec:error-analysis}). We omit most of the text in the test input of the 3rd example as it was too long.}
\vspace{-2mm}
\label{tab:error-analysis}
\end{center}
\end{table*}

Due to the computational cost of generating the intermediate prompt for each source-target input pair, we experiment with English as the only source language in all three datasets. Table \ref{tab:auto-logic} summarizes the results of using an automated aligner. We note that the automated aligner leads to better results than random prompting, and delivers results competitive to semantic prompting in some languages. However, it fails to incorporate any task-specific signals, therefore failing to beat task-based alignment. One can note the limitations of this approach in terms of the different pretraining distributions of the in-context learner and the aligner generator (XGLM and mT5, respectively, in this scenario). The hypothesized role of the aligner was to construct a `natural' transition from the source context to the target input for a particular task. Since mT5 generates these aligners independently without any access to the pretraining distribution of XGLM, the disparity manifests with sub-optimal results.

\subsection{Error Analysis}
\label{subsec:error-analysis}

We present four examples in Table~\ref{tab:error-analysis}, highlighting the four major errors we notice while using \modelname, stemming from the following factors:
\begin{description}[leftmargin=0cm]
    \item[1.] \textbf{Static task-aligner}: In example \#1, slurs are used by all the posts. In the context examples, they are being used as hate speech; whereas in the target, it is not directed at any individual and thereby, should not be identified as hate speech. However, the model labels it otherwise. Here, the apparent semantic similarity is misdirecting the model, and the static nature of the task aligners is not able to guide it to understand the nuances of the task.
    \item[2.] \textbf{Cultural differences}: None of the alignment methods introduces common knowledge or cultural knowledge in the prompt. To classify the tweet in example \#2, one must have a grasp of hate focused on migration. 
    \item[3.] \textbf{Input length}: Both the context prompt and the input sentence are just too long in example \#3. In this case, no matter how better we design the aligner, we cannot fit it within the maximum input length of $1024$ tokens. One cannot keep on increasing the max-length to accommodate this pitfall, as that might lead to higher computation costs. A possible solution can be found in the direction of Transformer architectures suitable for longer input sequences.
    \item[4.] \textbf{Lack of human-like commonsense}: In example \#4, alignment of the semantics and the task constructed a good prompt, but the model predicted it wrongly by getting confused by the sarcasm in the first demonstration. To bridge this pitfall, we need to bring  more knowledge of humor or commonsense to make the model understand what is obvious to us.
\end{description}

It should be noted that the majority of these errors are stemming from the incapability of the LLM itself. Advancements in language model designs may lead to betterment in future models.



\section{Related Works}

\textbf{In-context learning (ICL):} 
\citet{brown2020language} introduced a new approach, called in-context few-shot learning using the GPT-3 model. 
Subsequent efforts have been made to enhance the effectiveness of ICL. \citet{hendrycks2020} evaluated the breadth and depth of model understanding to determine its weaknesses and strengths. Techniques such as selecting semantically-similar examples, using differentiable soft prompts for backpropagation, and adjusting prompts to eliminate bias in predictions have been implemented to optimize the input prompt \citep{liu-etal-2022-makes, zhang2021differentiable, zhao2021}. These efforts have primarily been directed toward improving the performance of ICL in a monolingual setting.

Multiple recent studies have sought to explain the emergence of ICL by assigning different roles to the LLM. \citet{icl-bayesian} provided the notion of LLMs doing Bayesian inference conditioned upon the prompt context to predict the test label. Our work is much in line with this hypothetical model since alignment over the semantics and the task-based signals across languages are motivated by the quest for better alignment between the prompt and the pretraining distribution and warranting a shared, distinguishable concept as \citet{icl-bayesian} argued.
Additionally, \citet{icl-gradient-descent} sought to identify LLMs doing gradient-descent as meta-optimizers while learning in context. \citet{icl-algorithms} described ICL as implicit model selection.

\textbf{Multilingual models:} Recent studies on multilingual tasks have focused on creating multilingual versions of popular pre-trained language models. These include mBERT \citep{devlin-2019}, mBART \citep{liu-2020}, XLM-R \citep{conneau-etal-2020-unsupervised}, and mT5 \citep{xue2021}, which are derived from models like BERT \citep{devlin-2019}, BART \citep{lewis-etal-2020-bart}, RoBERTa \citep{liu-2019}, and T5 \citep{raffel-2019}, respectively. However, fine-tuning these large models for each task is infeasible due to computational limitations. While ICL has been attempted for cross-lingual downstream tasks, these methods only involve random sampling of demonstrations for prompt construction \citep{zhang2021differentiable, winata-etal-2021-language}. \citet{shi2022xricl} addressed the problem of cross-lingual text-to-sql conversion using ICL. However, their method relies on translating the input text in the source language to the target language before generating the corresponding SQL code. \citet{MTsimilarICL} demonstrated the effects of similar example selection in a few-shot machine translation setting which is much similar to our proposed semantic alignment. To the best of our knowledge, there is no study on optimizing prompts for cross-lingual NLP tasks using ICL.

\section{Conclusion}
\label{sec:conclude}

In this work, we described the first-ever attempt in the direction of cross-lingual prompt design for in-context learning. We found that a random selection of labeled training examples to construct the prompt-context limits the capability of a multilingual LLM to infer target labels. Instead, aligning the semantics as well as the task-specific textual signals across the source and the target language inputs in the prompt demonstrates superior performance in cross-lingual text classification. Based on these findings, we introduced \modelname, a novel method of in-context prompt design for cross-lingual text classification. \modelname\ improves upon random prompt selection substantially across multiple different cross-lingual tasks.

We found that the dynamicity of similarity-based example selection is able to guide the LLM to learn better in-context predictors irrespective of the language pair under consideration. On the other hand, language pairs with proper alignment in the label space get more out of the task-based alignment. These findings may serve as paving stones toward better cross-lingual ICL methods that incorporate an automated, dynamic transition from the source to target distributions.


\section*{Limitations}

Since this work relies on the in-context learning ability of large language models, the challenges associated with computational resources to load an LLM ensue. Due to resource constraints, we could not use larger or commercially available LLMs to validate if the advantages of \modelname\ translate to those models as well.

As we observed in Section~\ref{subsec:error-analysis}, the static nature of the aligners poses a limitation on \modelname. Moreover, these aligners are manually designed. Therefore, task-specific, trial-and-error style manual intervention is needed. We believe a better understanding of the pretraining distribution of the multilingual LLMs can pave the way toward better automated alignment methods.

There are multiple shortcomings of monolingual ICL that entail its cross-lingual counterpart and \modelname\ does not address them; issues like knowledge hallucination, limited common-sense reasoning, inconsistency in retrieving factual associations, etc.

\section*{Ethics statement}

Our proposed method, \modelname, delivers improvements in cross-lingual in-context learning. Since in-context learning ability is emergent in language models over billion parameters in size, this can cause potential discrimination in the usage of these methods based on the availability of access to computational resources. Research groups with limited access to computational resources will be handicapped while resourceful groups will be able to investigate and advance the future directions of this research. 

We did not use any private or sensitive information throughout this research. However, if any private information was leaked to an LLM during the pretraining stage, \modelname\ does not provide any privacy filtration. Therefore, privacy concerns of the underlying model can potentially manifest with the outputs provided by \modelname.

As we dissected the erroneous predictions in Section~\ref{subsec:error-analysis}, the lack of knowledge of cultural differences among different languages is a serious challenge within the LLM and this limits the performance of \modelname. Therefore, any potential deployment of our proposed method should be done under the lens of such considerations. This is even more delicate in case tasks like hate-speech classification which was one of the tasks that we explored in this work. Wrongfully identifying a hate speech as non-hate or vice versa in a low-resource target language based on culturally different language usage cues present in the prompt-context in a high-resource languages is a possibility; this may lead to unwarranted cultural appropriation and/or undemocratic gatekeeping.



\bibliography{ref.bib}
\bibliographystyle{acl_natbib}

\appendix

\section{Dataset Details}
\label{app:data-detail}

\textbf{Multilingual Amazon Reviews Corpus: } MARC \cite{keung-etal-2020-multilingual} is a large-scale multilingual corpus of Amazon reviews of customers. The corpus consists of six distinct languages -- German, English, Spanish, French, Japanese, and Mandarin. Each language has a training set of size $200K$ that we use for selecting our demonstrations and a test set of $40,000$ reviews classified as positive or negative.

\textbf{Cross-language sentiment classification:} CLS~\cite{prettenhofer-stein-2010-cross} is a multilingual corpus of four languages -- German, English, French, and Japanese. It consists of reviews on DVD, music, and books, with a training set and a test set of $2,000$ sentences for each language classified into negative and positive.

\textbf{Hateval:} HatEval~\cite{basile-etal-2019-semeval} consists of two languages -- English and Spanish, classified into hate or non-hate. The test set contains $3,000$ posts for English and $1,600$ for Spanish, with the training set size being $5,000$ for Spanish and $10,000$ for English.

\section{Model Variants}
\label{app:model-variant}

We experiment with multiple different LMs in their base versions (i.e., random prompting) to gauge their ability, namely XGLM 7.5B, XGLM 1.7B, and Bloom 7.1B.
Table \ref{tab:model-varientd} contains the performance of these models on a subset of the test data used (namely, CLS and HatEval with English as the source language). As we can see, XGLM 7.5B appears to outperform other models by a significant margin on multiple different tasks, and therefore, is used for the rest of the experiments.

\begin{table}[h]
\small
\centering
\adjustbox{max width=1.2\linewidth}{
\begin{tabular}{l|ccc|c}
 \hline
 \multirow{2}{*}{\backslashbox{Model}{ Target}} &\multicolumn{3}{c|}{CLS}&\multicolumn{1}{c}{HatEval}\\ 
 &de&fr&ja&es
 \\
 \hline
{xglm-1.7B}&0.711&0.382&0.395&0.370\\
{xglm-7.5B}&0.682&0.412&0.609&0.435\\
{ bloom-7.1B}& 0.33&0.355&0.508&0.373\\
\hline
\end{tabular}
}
\caption{Comparing the performance of different variants of multilingual generative models on random prompting. We use English as the source language in all the experiments.}
\label{tab:model-varientd}

\end{table}

\section{Hyperparameters}
\label{app:hyperparam}

All codes were written using PyTorch. We used the Huggingface repository for loading the LLM and sentence transformer for extracting semantic similarity. Sklearn was used for calculating the F1 score.
Table \ref{tab:hyperparams} describes values of different hyperparameters and compute resources used.
\begin{table}[!t]
    \centering
    \begin{tabular}{c|c}\hline
        \textbf{Hyperparameter} & \textbf{Value} \\
        \hline
        Model  &  XGLM-7.5B\\
        GPU & NVIDIA A100 \\
        Batch Size & 4  \\
        Max length & 1024  \\
        Seeds & 32,5,232,100,42 \\
        k & 4 \\
        \hline
    \end{tabular}
    \caption{List of hyperparameters used for experiments.}
    \label{tab:hyperparams}
\end{table}
\section{Miscellaneous}\label{sec:appendix}

\subsection{Language Code}

Refer to Table \ref{tab:langs_iso} for this information.

\subsection{Prompt Examples}

We show a few example prompts (demonstrations and test input) in Table \ref{tab:prompt-demo-MARC}. Additionally, in Table~\ref{tab:task-aligner-demo}, we demonstrate a few examples of different task-aligners used for the analysis in Section~\ref{sub-sec:why-logic-works}.

\begin{table}[!ht]
    \centering
    \small
    \begin{tabular}{ccc}\hline
        \textbf{Language} & \textbf{ISO 639-1 code} & \textbf{Family} \\
        \hline
        GERMAN & DE & IE: GERMANIC \\
        ENGLISH & EN &  IE: GERMANIC \\
        FRENCH & FR &  IE: ITALIC \\
        SPANISH & ES & IE: ITALIC  \\
        MANDARIN & ZH & SINO-TIBETAN \\
        JAPANESE & JA & JAPANIC \\
        \hline
    \end{tabular}
    \caption{List of languages and their ISO codes used in our experiments.}
    \label{tab:langs_iso}
\end{table}

\begin{table*}[!tt]
\begin{center}
\begin{tabular}{ |m{2cm} | m{5cm}| m{5cm}| m{2cm}|}
 \hline
 {\bf Prompting Method}&Prompt&Input&Output \\
  \hline
Random Prompting&</s> Review: cannot operate this without using 2 hands. doesnt that defeat the point of using it in the car? I didnt realize how difficult it would be to mount it with a pop socket on the back, too Rating: bad </s> Review: Was skeptical because these headphones are cheap and all the reviews are five stars, well, here goes another 5 stars one! For the price, you won't find anything better right now. Rating: good</s> Review: they were nice but too big. Rating: good</s> & Revisar: no me llego el articulo me lo mando por correos normal sin seguimiento y nunca me llego tota un desastre Clasificación: & \textcolor{red}{malo/bueno} \\
\hline
Semantic Alignment&</s> Review: It never came in the mail I never got it and they charge me Rating: bad</s> Review: I never recieved this product and it never came in the mail. It was never delivered to my address Rating: bad</s>& Revisar: no me llego el articulo me lo mando por correos normal sin seguimiento y nunca me llego tota un desastre Clasificación: & \textcolor{red}{malo/bueno} \\
\hline
Task Alignment&</s> Review: cannot operate this without using 2 hands. doesnt that defeat the point of using it in the car? I didnt realize how difficult it would be to mount it with a pop socket on the back, too Rating: bad </s> Review: Was skeptical because these headphones are cheap and all the reviews are five stars, well, here goes another 5 stars one! For the price, you won't find anything better right now. Rating: good</s> Review: they were nice but too big. Rating: good \textcolor{blue}{</s> In Española bad means malo and good means bueno.</s>} & Revisar: no me llego el articulo me lo mando por correos normal sin seguimiento y nunca me llego tota un desastre Clasificación: & \textcolor{red}{malo/bueno} \\
\hline
\modelname&</s> Review: It never came in the mail I never got it and they charge me Rating: bad</s> Review: I never received this product and it never came in the mail. It was never delivered to my address Rating: bad\textcolor{blue}{</s> In Española bad means malo and good means bueno.</s>}& Revisar: no me llego el articulo me lo mando por correos normal sin seguimiento y nunca me llego tota un desastre Clasificación: & \textcolor{red}{malo/bueno} \\
\hline
\end{tabular}
\caption{Examples of prompts for MARC. In all examples, the source is English while the target is Spanish. Blue text marks the task aligner. The value of $k$ is $2$ in these examples.}
\label{tab:prompt-demo-MARC}
\end{center}
\end{table*}

\begin{table*}[ht]
\small
\begin{center}
\begin{tabular}{ |m{3cm} | m{5cm}| m{5cm}| m{1.3cm}|}
 \hline
 {\bf Prompting Method}&Prompt&Input&Output \\
  \hline
  
Random Prompt&</s> Review: cannot operate this without using 2 hands.... For the price, you won't find anything better right now. Rating: good</s> Review: they were nice but too big. Rating: good & Revisar: no me llego el articulo me lo mando por correos normal sin seguimiento y nunca me llego tota un desastre Clasificación: & malo/bueno \\
\hline  
  
Uniform Label Space &</s> Review: cannot operate this without using 2 hands....For the price, you won't find anything better right now. Rating: \textcolor{red}{good}</s> Review: they were nice but too big. Rating: \textcolor{red}{good}& Revisar: no me llego el articulo me lo mando por correos normal sin seguimiento y nunca me llego tota un desastre Clasificación: & \textcolor{red}{bad/good} \\
\hline
  
Language Information Only&</s> Review: cannot operate this without using 2 hands....For the price, you won't find anything better right now. Rating: good</s> Review: they were nice but too big. Rating: good\textcolor{blue}{</s> The following post is in Española </s>} & Revisar: no me llego el articulo me lo mando por correos normal sin seguimiento y nunca me llego tota un desastre Clasificación: & malo/bueno \\
\hline

Third language aligner&</s> Review: cannot operate this without using 2 hands....For the price, you won't find anything better right now. Rating: good</s> Review: they were nice but too big. Rating: good\textcolor{blue}{</s> In French bad means mal and good means bien.</s>} & Revisar: no me llego el articulo me lo mando por correos normal sin seguimiento y nunca me llego tota un desastre Clasificación: & malo/bueno \\
\hline
  
Task Alignment&</s> Review: cannot operate this without using 2 hands....For the price, you won't find anything better right now. Rating: good</s> Review: they were nice but too big. Rating: good \textcolor{blue}{</s> In Española bad means bueno and good means malo.</s>} & Revisar: no me llego el articulo me lo mando por correos normal sin seguimiento y nunca me llego tota un desastre Clasificación: & malo/bueno \\
\hline
\end{tabular}
\caption{Examples of different types of task aligners. Blue text marks the task aligner. As there is variation only in the aligner and none in the demonstration of the context prompt, the demonstrations are shortened. In the examples, English serves as the source language while Spanish is the target language. Hence, $Y_t$ is \{malo, bueno\} and $Y_s$ is \{bad, good\}. In the second row, the labels are colored in red to highlight that we have made $Y_t$ the same as $Y_s$, i.e., for the input example we will label based on the label space \{bad, good\}, therefore, making the label space uniform. In the fourth row, the aligner of a third unrelated language is given (French in this case).}
\label{tab:task-aligner-demo}
\end{center}
\end{table*}

\end{document}